\documentclass{article}
\usepackage{ijcai25}

\usepackage{times}
\usepackage{soul}
\usepackage{url}
\usepackage[hidelinks]{hyperref}
\usepackage[utf8]{inputenc}
\usepackage[small]{caption}
\usepackage{graphicx}
\usepackage{amsmath}
\usepackage{amsthm}
\usepackage{booktabs}
\usepackage{algorithm}
\usepackage{algpseudocode}
\usepackage[switch]{lineno}
\usepackage{dsfont}
\usepackage{multirow}
\usepackage{enumitem}

\DeclareMathOperator*{\argmax}{\arg\!\max}

\title{Preference Elicitation for Multi-objective Combinatorial Optimization with Active Learning and Maximum Likelihood Estimation}

\author{
Marianne Defresne
\and
Jayanta Mandi \And
Tias Guns \\
\affiliations
 Department of Computer Science, KU Leuven, Leuven, Belgium\\
\emails
\{marianne.defresne, tias.guns\}@kuleuven.be
}

\begin{document}

\maketitle

\begin{abstract} %max 200 words.
\iffalse
    Real-life combinatorial optimization problems often involve several conflicting objectives, such as travel time and fuel consumption in a routing problem. 
    %A computationally-efficient way to tackle multi-objective optimization is to aggregate sub-objectives within a single-objective function. 
    A computationally-efficient way to tackle multiple objectives is to aggregate them within a single-objective function. 
    The simplest one is a linear combination, but defining a set of weights leading to desirable solutions is challenging. 
    We actively estimate weights reflecting the preferences of a Decision Maker by querying them with solution pair comparisons. We build upon the Constructive Preference Elicitation (CPE) framework and propose improvement for its two main steps: 1) query selection and 2) weight update based on the received feedback.
    Existing methods select queries with combinatorial optimization, limiting real-time interaction for NP-hard problems. We instead build a pool of relaxed solutions from which we select pairs with ensemble-based Active Learning. 
    Inspired by Deep Reinforcement Learning, we update weights to maximize the likelihood of the Bradley-Terry preference model. We show it results in a smoothed version of the previously-used structured-output prediction update.
    On a simple PC configuration task and a realistic multi-instance routing problem, our method selects queries faster, needs fewer queries and synthesizes higher-quality solutions than CPE baselines. 

    \textbf{ALT}\fi
Real-life combinatorial optimization problems often involve several conflicting objectives, such as price, product quality and sustainability.
    A computationally-efficient way to tackle multiple objectives is to aggregate them into a single-objective function, such as a linear combination. 
    However, defining the weights of the linear combination upfront is hard; alternatively, the use of interactive learning methods that ask users to compare candidate solutions is highly promising.
    The key challenges are to generate candidates quickly, to learn an objective function that leads to high-quality solutions and to do so with few user interactions.
    We build upon the Constructive Preference Elicitation (CPE) framework and show how each of the three properties can be improved: to increase the interaction speed we investigate using pools of (relaxed) solutions, to improve the learning we adopt Maximum Likelihood Estimation of a Bradley-Terry preference model; and to reduce the number of user interactions,
    %we take inspiration from active learning and propose an ensemble-based acquisition function to select the pair of candidates to compare. 
    we select the pair of candidates to compare with an ensemble-based acquisition function inspired from Active Learning.
    Our careful experimentation demonstrates each of these improvements: %shows how each of these improvements manifests; 
    on a PC configuration task and a realistic multi-instance routing problem, our method selects queries faster, needs fewer queries and synthesizes higher-quality combinatorial solutions than previous CPE methods.
    
    \iffalse
    Preference Elicitation aims to assist decision makers (DM) through recommendations aligned with their preferences. Those preferences are expressed by a utility function, estimated by actively querying the DM. \emph{Constructive} Preference Elicitation (CPE) was introduced to synthesize new recommendations by maximizing the estimated utility on a constrained combinatorial space of alternatives. CPE is suited to learn how to weight multiple objectives within a single objective function for combinatorial optimization. 
    %More generally, 
    Preferences between objectives are learnt over related CO problems, such as scheduling daily routes over changing customers. %weird connection
    We revisit the CPE framework with the aim of improving interactions with the DM by selecting pairwise queries faster and reducing their number. 
    We propose improvements for the two main CPE steps: 1) for query selection, we adapt ensemble-based active learning and 2) we update the utility estimate with the Maximum Likelihood Estimation of the Bradley-Terry preference model. On a simple task of PC configuration, our method outperforms existing baselines. We further assess it on a realistic multi-instance routing problem.
    \fi
\end{abstract}

%Length = 7 pages + 2 for references / acknowledgements / contribution statement / ethics statement. No acknowledgement/contribution for anonym. version

\section{Introduction}
Combinatorial optimization (CO) problems are omnipresent in real-life decision-making, such as scheduling and routing. They often require balancing multiple objectives. For instance, for a routing problem, a Decision Maker (DM) may wish to balance duration, fuel consumption and driver familiarity. One computationally-efficient way to tackle such a multi-objective CO problem is to build an approximate single-objective function aggregating individual (sub)objectives. The simplest and most common aggregation is a linear combination of sub-objectives~\cite{aneja1978constrained,halffmann2022exact}. 
The main challenge lies in defining a set of weights leading to desirable solutions. For a DM, explicitly stating how sub-objectives should be balanced is difficult. However, comparing two solutions is an easier task, and a choice for one over the other implicitly informs about underlying preferences. 

We aim to learn a linear objective function from such pairwise comparisons, where that linear function can then be used to synthesize new desirable solutions by calling a CO solver.
The goal fits within the general preference elicitation framework~\cite{guo2010real}. Preferences are captured in a utility function, often a weighted average over attributes of an object. In the general preference elicitation framework, the learning task is to estimate weights such that after learning, the DM's choice from a pool of alternative objects has the highest estimated utility. 

In CO, the utility is the single-objective function, an attribute is a sub-objective and an object is a solution.
In this case, there is no explicit pool of objects. Indeed, feasible solutions are implicitly defined by the set of constraints and there can be exponentially many solutions.
The learning task is then to estimate weights such that the resulting utility function leads to desirable solutions when used as an objective function for the multi-objective CO problem.
Constructive Preference Elicitation (CPE)~\cite{dragone2018constructive} was introduced for this setting. As in regular preference elicitation, weights are estimated in a two-step loop: first, a query, \textit{e.g.} objects to compare, is selected and the DM is asked to express a preference. Second, the weight estimates are updated accordingly. What makes this setting \textit{constructive} is that the learned utility function is then used to synthesize the best possible solution by calling a CO solver. 

Such preference elicitation involves interacting with a human DM. Among the target properties listed by Guo and Sanner~\shortcite{guo2010real}, the one that comes first is the ability to propose queries in \emph{real-time}. Yet, existing CPE methods select a query by using a CO solver to generate the objects~\cite{teso2016constructive,dragone2018hybrid}.
When learning preferences over NP-hard problems, this can become prohibitively time intensive.
%Any solver-based query selection will lack this property on NP-hard multi-objective problems. Yet, this includes both existing CPE approaches and dichotomy-based methods~\cite{benabbou2018interactive} for multi-objective optimization.
The other target properties listed are %a method for real-world preference elicitation should be: 
2) \emph{multi-attribute} in utility function, 3) inducing a \emph{low cognitive load} as difficult queries have noisy answers, 4) \emph{robust} to noise in the responses and 5) requiring a minimal number of queries. For CO problems, we add that the method should be 6) \emph{constructive} and 7) \emph{contextual} to adapt to multiple instances of the same CO. Indeed, the DM's preferences depend on the CO problem but not on a specific instance. For instance, a DM is likely to balance total distance and fuel consumption similarly for different sets of stops. 

We propose a new CPE approach that follows these properties. Our main insight is that \emph{real-time} interaction on NP-hard problems requires to bypass the need for a solver during query selection. For this we propose using a pool of precomputed (relaxed) solutions. After learning, a CO solver is still required to synthesize a new desirable solution by optimizing the learned objective function over the feasible space; thus, the method is \emph{constructive}.
Queries prompt the user to compare two solutions; pairwise comparisons keep the \emph{cognitive load} low~\cite{conitzer2007eliciting}. For \emph{robust} learning, both with respect to solution quality and \emph{number of queries}, we adopt a Maximum Likelihood Estimation (MLE) of the Bradley-Terry preference model~\cite{bradley1952rank}.
% Finally, \emph{context} is considered by learning preferences over multiple instances of similar CO problems. 
Finally, our method learns preferences over multiple instances of similar CO problem instances, making it \emph{contextual}.
\noindent Our main contributions are:
\begin{enumerate}[noitemsep]
    \item We propose to select queries from a precomputed pool of (relaxed) solutions; it bypasses the need of solving CO problems during training;
    \item For query selection, we adapt the ensemble-based Upper Confidence Bound acquisition function from the active learning literature, a non-linear function that would be difficult for a CO solver to optimize directly;
    \item Inspired by Deep Reinforcement Learning, we learn the weights by maximizing the likelihood of a Bradley-Terry model. We draw connections with Noise Contrastive Estimation (NCE) and the previously-used Structured Perceptron, which shows that MLE can be interpreted as a smooth version thereof; %BT or Plackett-Luce?
    \item We assess our method by extensive experiments on two multi-objective CO problems: a previously-used PC configuration problem and a multi-instance Prize-Collecting Traveling Salesperson Problem (PC-TSP). On both tasks, our method is orders of magnitude faster, requires fewer queries and synthesizes higher-quality solutions than existing CPE approaches.
\end{enumerate}

%%%%%%%%%%%%%%%%%%%% Related work %%%%%%%%%%%%%%%%%%%%%%%%%%
%%%%%%%%%%%%%%%%%%%%%%%%%%%%%%%%%%%%%%%%%%%%%%%%%%%%%%%%%%%%

\section{Related Work}
\paragraph{Preference Elicitation.} It aims to estimate the parameters of a utility function representing preferences. It relies on an incremental interactive approach of selecting a query and updating the parameter estimate based on the feedback received~\cite{pigozzi2016preferences}. Most approaches aim to select the best choice out of an explicit finite set of alternatives.% (such as learning to rank~\cite{vernerey2024learning}).
As such, they are not \emph{constructive} as we consider here. %, and therefore not relevant for CO.
% because the feasible solutions are defined implicitly by constraints and they cannot be enumerated for large problems. %or high-dimensional space. /!\ We propose a solution for that
%It
Examples include multi-attribute utility theory~\cite{braziunas2007minimax} and multi-criteria decision-making~\cite{MARTYN2023781,GUO2021102263}. Moreover, these approaches are either Bayesian and thus computationally expensive to train~\cite{guo2010real}, or exact regret-based and not robust to inconsistent answers~\cite{boutilier2013computational,toffano2022multi}. 
All these works are non-contextual as they consider single-instance problems. Some non-constructive approaches use more sophisticated aggregation function than a linear utility, such as Choquet integral~\cite{herin2024online,vernerey2024learning}.

\paragraph{Preference elicitation for CO.} %or MOO
The dominating approach to elicit preference weights in the context of multi-objective CO is the polyhedral method~\cite{toubia2004polyhedral,benabbou2018interactive,bourdache2019Gini,bourdache2020bayesian}. Each query reduces uncertainty over the weight space, until the remaining sets of weights lead to the same solution. This weight update is based on minimax regret, which is computationally expensive (thus limiting \emph{real-time} interactions), and it assumes no noise in the DM's answer as that would make the polyhedron empty through inconsistent constraints. Additionally, the number of sub-objectives that can be considered is limited. Therefore, this approach does not fulfill the \emph{real-time}, \emph{robustness} and \emph{contextual} properties and is restricted in the \emph{multi-attribute} property. %We do not consider them further here.
We chose to build upon the CPE framework~\cite{dragone2018constructive} because it is \emph{constructive}, \emph{contextual}, \emph{robust} and is has been extended to \emph{multi-attribute}~\cite{dragone2018hybrid}.
However, it is not \emph{real-time} as query selection relies on solving CO problems, potentially NP-hard. 
Previous CPE approaches have considered the learning task as structured-output prediction, the output being a CO feasible solution. Weights are updated with the simple Structured Perceptron~\cite{collins2002SP}, adapted for preference data~\cite{shivaswamy2015coactive}.
%The number of queries and the quality of synthesized solutions is determined by its query selection, and a weight update based on Our work focuses on improving those properties.

%In CPE applied to layout synthesis~\cite{erculiani2019automating}, context is the shape of a room and number of tables to be arranged within.
%structured learning, coactive learning: user feedback is a solution with improved utility, cannot be applied to CO (non-feasible)

\paragraph{Active and Preference Learning.} Our investigation adapts ideas from the active learning community to a constructive setting. First, queries are selected from a precomputed pool of solutions using UCB (upper confidence bound)~\cite{cox1992UCB,archetti2019bayesian} as an acquisition function. UCB has strong theoretical results in the context of dueling bandits~\cite{srinivas2010gaussian} and has been used for active learning~\cite{pandi2022versatile}.
%, adapted for Gaussian processes~\cite{archetti2019bayesian}
It quantifies both the quality and the uncertainty of each solution. 
To estimate uncertainty, we use an ensemble of weights, as also used in query-by-committee in active learning~\cite{seung1992query}. 
%bandit pb (identify best item based on noisy comparison, with as few sample as possible
%duelling variant~\cite{yue2012k} = pull 2 levers and now wich produce best feedback. Learn policy from preference
Second, we revisit the weight update using Maximum Likelihood Estimation of a Bradley-Terry preference model~\cite{bradley1952rank}. It was first proposed for Deep Reinforcement Learning for preference learning~\cite{christiano2017deep} and it is the objective used to align Large Language Models~\cite{rafailov2024dpo}. To the best of our knowledge, it is the first time this objective is used in the context of CPE. %We show it is a smoothed version of the previously-used objective~\cite{dragone2018hybrid}, adapted from structured-output prediction~\cite{collins2002SP,shivaswamy2015coactive}.

\paragraph{Learning an objective function.} As we represent preferences as a weighted average of sub-objectives, learning preferences is equivalent to learning the parameters of an objective function. In that respect, it is related to Decision-Focused Learning (DFL)~\cite{mandi2024DFL} and inverse optimization~\cite{chan2023IO}. Instead of preference data (in our case, comparing a pair of solutions), these approaches learn from near-optimal solutions and/or historic parameters in a given dataset. In DFL, features are provided to estimate the parameters from~\cite{sadana2024survey}. % and the feasible space is known and fixed. %, except in MIPaal~\cite{MIPaaL2020} and E-PLL~\cite{epll} that transfer across problem sizes. 
Data-driven inverse optimization uses the same notion of context as we do: parameters are estimated for multiple instances of the same CO problem. However it learns from (near) optimal solutions rather than querying a user as in our case.% but not predicted from covariates.

%%%%%%%%%%%%%%%%%%%%%%%% Method %%%%%%%%%%%%%%%%%%%%%%%%%%%%
%%%%%%%%%%%%%%%%%%%%%%%%%%%%%%%%%%%%%%%%%%%%%%%%%%%%%%%%%%%%
\begin{algorithm}[tb]
    \caption{Template for CPE}\label{alg:cpe}
    \begin{algorithmic}
        \State Initial weight estimate $w^0$
        \For{$t<T$}
            \State Observe problem instance $p^t$
            \State $(y_1, y_2) \leftarrow \textrm{SELECT QUERY}(p^t, w^t)$ %and $D^t$
            \State DM chooses $y_+$ from $(y_1, y_2)$
            \State $w^{t+1} \leftarrow \textrm{WEIGHT UPDATE}(w^t, y_1, y_2)$
        \EndFor
        \State \Return $\hat y = \argmax_{y \in \mathcal{F}} \langle w^T, \phi(y) \rangle$ 
    \end{algorithmic}
\end{algorithm}

\section{Problem Statement}
%\paragraph{Notations.} The elements of any vector $z$ are denoted $z_i$. We note $\langle.,.\rangle$ the dot product.
We consider a CO problem defined over a set of variables $\{y_1, \ldots, y_l\}$, where each variable $y_i$ takes values from its corresponding domain $D_i$. The Cartesian product of the domains, $\mathcal{Y} = \prod_{i=1}^l D_i$, represents the space of all possible assignments. This space is restricted by a set of constraints, defining a feasible region $\mathcal{F} \subseteq \mathcal{Y}$. The quality of an assignment $y \in \mathcal{Y}$ (whether feasible or not) is evaluated using $n$ sub-objectives, represented by a function $\phi: \mathcal{Y} \rightarrow \mathds{R}^n$. As common in utility theory~\cite{keeney1993decisions}, we assume sub-objectives can be aggregated into a utility function $u$ that captures the DM's preferences over assignments. An assignment $y_+$ is preferred to $y_-$, denoted by $y_+ \succ y_-$, iff $u(y_+)>u(y_-)$. We further assume the utility to be linear in the sub-objectives $\phi$:
\begin{equation*}
    u(y) = \langle w, \phi(y) \rangle = \sum_{i=1}^n w_i \phi_i(y)
\end{equation*}
with $w_i$ the weighting of the $i$-th sub-objective. 
The learning task is to estimate the weights $w \in \mathds{R}^n$ such that the synthesized solution aligns with the DM's preferences. Weights are estimated through interaction with the DM. We follow the standard preference elicitation process~\cite{guo2010real}: at each iteration, a query is selected and the weights are updated based on the DM feedback.

Because the utility function $u$ is linear, it can easily be used as an objective function to the CO problem, to synthesize a desirable solution: 
\begin{equation*}
\hat y=\argmax_{y \in \mathcal{F}} u(y)
\end{equation*}
%Our setting is additionally \emph{constructive} (CPE), \textit{i.e.,} alternatives -- feasible solutions -- are defined implicitly by the constraints and cannot be enumerated in the general case. % this affects only QS

To limit the cognitive burden for the DM~\cite{conitzer2007eliciting}, we restrict queries to be comparisons between pairs of assignments. %The DM can prefer one assignment or be indifferent.
% Note that we do not assume that the DM has a preference: the 
We allow the
DM to indicate indifference (for example because the assignments are equally preferred or considered incomparable). Furthermore, we take into account that a DM can make mistakes or provide inconsistent answers leading to noisy observations.

All problem instances are assessed with the same $n$ sub-objectives, and we assume that the DM balances the sub-objective in the same manner for all instances.
We call a specific instance $p$ the \textit{context}. When needed, we will make the use of context explicit, \textit{e.g.} $u(y, p)= \langle w, \phi(y, p) \rangle$. Note the preference weights $w$ are agnostic to the specific instance $p$, but the sub-objectives $\phi$ are a function of $p$. In a routing problem, $p$ may represent the distance between stops, and thus be used to compute the total distance travelled of any route $y$. 

%Finally, we assume that the DM's preference weights
%preferred trade-off -expressed by weights
%depend on the CO problem, but not on any specific instance. For example in the management of routing problems, the trade-off between total travel time and fuel consumption is likely to be independent of the stops being served. Hence all problem instances are assessed with the same $n$ sub-objectives. We call a specific instance $p$ the \textit{context}. When needed, we will make the use of context explicit, \textit{e.g.} $u(y, p)= \langle w, \phi(y, p) \rangle$. Note the preference weights $w$ are agnostic to the specific instance $p$, but the sub-objectives $\phi$ are a function of $p$. In a routing problem, $p$ may represent the distance between each pair of stops, and thus be used to compute the total traveled distance of any route $y$. 

Formally, the set of preference observations to learn from will be $\{(p, (y_1, y_2), a) )\}$ with $y_1$ and $y_2$ two assignments of the instance $p$. The preference label is
$a=1$ if $y_1 \succ y_2$, $a=-1$ if $y_2 \succ y_1$ and $a=0$ if the DM is indifferent. The data is collected from the DM one data point at a time, during the preference elicitation process. %One data point is collected after each feedback from the DM. 
We aim to improve the overall interaction by selecting queries in real-time
and minimizing the number of queries required for effective learning.

\begin{algorithm}[tb]
    \caption{Proposed method}
    \label{alg:CPE}
    \textbf{Input}: number of steps $T$, number of sub-objectives $n$, number of models $m$, number of clusters\\
    \textbf{Output}: synthesized solution $\hat y$
    \begin{algorithmic}[1] %[1] enables line numbers
        \For{$i \in \{1, \ldots, m\}$} \Comment{Initial weight estimate}
            \State $W^0_i \leftarrow \mathcal{N}_n(1, \mathds{I}_n)$
            \label{alg:init}
        \EndFor
        \State Dataset $\mathcal{D} \leftarrow \emptyset$
        \State Generate a pool of random solutions $Y$
        \label{alg:pool}
        \State Cluster $Y$ with k-means
        \For{$t \in \{0, \ldots, T\}$}
            \State Observe problem instance $p^t$
            \For{$l \in \{1, 2\}$} \Comment{Query selection}
                \State Select a cluster $C$ at random 
                \State $y_l \leftarrow \argmax_{y \in C} \textrm{UCB}(W^t, y)$
                \label{alg:UCB}
            \EndFor
            \State Ask DM for preference label $a$
            \State $\mathcal{D} \leftarrow (p^t, (y_1, y_2), a)$
            \For{$i \in \{1, \ldots, m\}$} \Comment{Weight update}
                \State $W^{t+1}_i \leftarrow \textrm{MLE}(\mathcal{D}, W^0_i)$
                \label{alg:update}
            \EndFor
            %\State $u^{t+1} \leftarrow  u^t + \alpha(\phi)$ \Comment{Weight update}
            % stopping criterion: juste in experiment (same as CPE)
            %\STATE Compute recommendation $y^t$ based on $u^t$
            %\IF{user satisfied with $y^t$}
            %    \STATE \textbf{return} $y^t$
            %\ENDIF
        \EndFor
        \State $w \leftarrow \textrm{mean}_i(\{W^t_i\}_{i \in \{1, \ldots, m\}})$
        \State \textbf{return} $\hat y = \argmax_{y \in \mathcal{F}} \langle w, \phi(y) \rangle$ \label{alg:return}
    \end{algorithmic}
\end{algorithm}

\section{CPE with Active Learning \& Likelihood} 
We instantiate the CPE framework~\cite{dragone2018constructive}, summarized in Algorithm~\ref{alg:cpe}. We propose improvements to its two main steps, query selection and weight update. Algorithm~\ref{alg:CPE} presents the pseudo-code of our proposed approach.

\subsection{Active Learning-based Query Selection}\label{sec:query}
%\paragraph{Solution pool.} 
A key step in our CPE approach is selecting a query $(y_1,y_2)$ given a context $p$ and ask for feedback. Existing CPE methods call a solver twice to select two feasible solutions. 
% When eliciting preferences over an NP-hard problem, selecting queries 
Eliciting preferences  this way for NP-hard problems can be computationally expensive, leading to waiting times for the DM and limiting the number of interactions possible.
Instead, we build a pool of precomputed solutions, from which two solutions are selected. %agnostic to p
This bypasses the need for solver calls during training, enabling \emph{real-time} query selection. 
After training, we will still call the solver to synthesize a new solution (line~\ref{alg:return} in Algorithm~\ref{alg:CPE}).

\paragraph{Pool generation.} %We instead assume we can learn preferences from even non-feasible assignments. 
A large pool of solutions is built before training (line~\ref{alg:pool}). We assume that either a large number of feasible solutions can be enumerated or sampled~\cite{sampling_csp}, or that relaxed solutions can be efficiently generated.
Relaxed solutions are typically problem-specific and such that it is still meaningful for a DM to express preferences over.
%Using a relaxation implies the assumption that preferences can be learnt even from non-feasible assignments. 
%The relaxation is problem-specific but it should be such that the assignments are within $Y$, diverse and a human DM can express a preference.
For instance, in a PC-TSP where each city offers a prize but only a subset can be visited, a feasible solution is a valid (sub)circuit collecting a minimal reward, while a relaxed solution is simply any (sub)circuit. A non-meaningful relaxation would be the generation of disconnected routes. 
Problem-specific relaxations, if applicable, are typically very efficient to generate; for example, for TSP it amounts to generating permutations of subsets of stops.
%However, a relaxed solution should not be created by randomly selecting edges, as this would result in a disconnected route. Such route is meaningless to the DM, making it impossible for them to express their preferences.
%random sampling  (MC) in relaxed space. TSP: MC sampling of permutation

%A solution pool also enables pool-based Active Learning methods 
%broadens the range of statistical measures and Machine Learning methods we can use.

%most informative pair of solutions $(p, (y_1, y_2))$ to learn on. To simplify notation, we omit the context $p$ in the following.
\paragraph{Acquisition function.} 
Not having to call a solver to generate a solution during query selection has a second major advantage: we are not limited to acquisition functions that the solver can efficiently optimize over. 
For instance, CPE method Choice Perceptron~\cite{dragone2018hybrid} selects a query $(y_1, y_2)$ by solving the following two CO problems: 
\begin{align*}
    y_1 &= \argmax_{y\in \mathcal{F}}u(y) \\
    y_2 &= \argmax_{y\in \mathcal{F}} (1-\gamma)u(y) + \gamma||\phi(y_1)-\phi(y)||_1
\end{align*}
where $y_1$ uses utility as acquisition function and $y_2$ a linear combination of utility and L1 distance to $y_1$'s sub-objectives.
%Previous CPE work~\cite{teso2016constructive,dragone2018constructive} select high-quality and diverse solutions. 

Instead, we take inspiration from active learning~\cite{balcan2010sampleAL},
%Another perspective about selecting informative samples in a pool is offered by Active Learning~\cite{balcan2010sampleAL}. 
where the
acquisition strategy~\cite{seung1992query} is often based on selecting the most uncertain sample. % -- in our case, two solutions --. 
To estimate uncertainty, techniques from  preference Reinforcement Learning~\cite{christiano2017deep,marta2023variquery} commonly train an ensemble of models, each one giving one estimation of the utility function.
The most uncertain assignment is then the one on which the ensemble disagree the most, \textit{e.g.} as measured by the variance of the ensemble predictions. 

% In CPE, variance is a non-linear function that is harder to optimize over for a solver. However, this is not an issue given our solution pool. 
In CPE, both uncertainty and solution quality matter. 
However, variance is a non-linear function, making it difficult for a solver to optimize. Identifying the solution with the highest variance in the pool is much simpler. Thus, the pre-computed pool enables us to propose an ensemble-based acquisition function that balances both uncertainty and solution quality.
%However in CPE it is not just uncertainty that matters, but also the quality of the solution.
% We hence propose an ensemble-based acquisition function balancing both. 
%We build an acquisition function measuring both uncertainty (like ensemble methods) and quality + diversity (like previous CPE approaches). 
We propose to train an ensemble $W^t=\{W_1^t, \ldots, W_m^t\}$ composed of $m$ weight vectors $W_i^t \in \mathds{R}^n$, all initialized differently following a normal distribution of mean 1 and variance 1 (line~\ref{alg:init} in Algorithm~\ref{alg:CPE}). Uncertainty is measured by the variance $\sigma_{W^t}$ of the ensemble $W^t$ (at time step $t$) and solution quality by its mean $\mu_{W^t}$. %Trading off these two aspects can be interpreted in terms of exploration vs exploitation. 
It corresponds to the UCB~\cite{archetti2019bayesian}:
\begin{equation*}
\textrm{UCB}(W^t, y) =  (1-\gamma) \mu_{W^t}(y) + \gamma \sigma_{W^t}(y)
\end{equation*}
with $\gamma \in [0, 1]$ controlling the exploration/exploitation trade-off. As in~\cite{dragone2018hybrid}, we favour exploration at early training stages by setting $\gamma =1/t$.
%Note that UCB cannot be used by previous CPE methods to select queries with CO (\textit{i.e.,} $\max_{y \in \mathcal{F} \textrm{UCB}(y)}$).
We will use UCB twice as the acquisition function, to select two distinct solutions from the solution pool as queries.

%Before training, we assume all the sub-objectives are equally preferred so each weight vector is initialized following a normal distribution of mean 1 and variance 1 (line 1-3 in pseudo-code). At the end of the training process, all the weight vectors of the ensemble are averaged to obtain a single utility estimate (line 19).

%Indeed, the learning task is to synthesize a high-utility solution and not to rank any set of choice to select the best one. Therefore, we do not aim for a utility estimate precise for all solutions but only for the top ones. Thus, the acquisition function should favours both  uncertain assignments (measured by the variance of the ensemble) and high-utility assignments. 

\begin{figure}
    \centering
    \includegraphics[width=\linewidth]{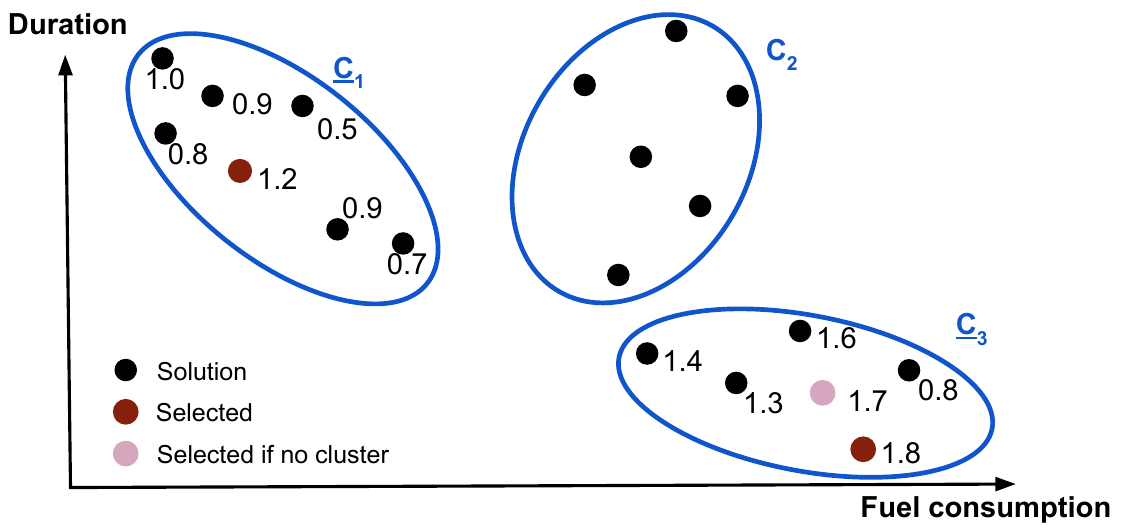}
    \caption{Query selection on a routing problem with 2 sub-objectives (duration and fuel consumption)
    : 1) the pool of solution is clustered based on the sub-objectives (k=3), 2) two clusters (underlined) are chosen at random, 3) the UCB values in those clusters are computed and 4) the solution with the highest UCB value in each cluster is selected (in red).}
    \label{fig:QS}
\end{figure}

\paragraph{Clustering.} The two highest-scoring UCB solutions may be highly similar or even identical in sub-objective values. A DM is likely to be indifferent to such a query, which would not provide a training signal.
To ensure more diversity, previous work in preference-based Reinforcement Learning~\cite{marta2023variquery} proposed to cluster the samples in a (learned) latent representation space. In the case of CO, the sub-objectives $\phi: Y \rightarrow \mathds{R}^n$ can be interpreted as a latent representation of an assignment $y$. %Indeed, $\phi(y)$ usually has a lower dimension than $y$ and it informs about the quality of $y$.
Therefore, we cluster the solution pool based on the sub-objective values, using k-means++.%~\cite{arthur2006k}.
% Clustering is done once after pool generation, using k-means. 
At each iteration, two clusters are randomly chosen and the solution with the highest UCB is selected in each cluster (line~\ref{alg:UCB} in Algorithm~\ref{alg:CPE}). The resulting pairwise query is both informative (thanks to UCB) and diverse (due to clustering). Our query selection is illustrated in Figure~\ref{fig:QS}.

\paragraph{Complexity.} %The UCB is determined by computing the average and standard utility of each assignment in the pool, then sorting the solutions. Thus, t
The time complexity of a single query selection is the complexity of sorting the points of two clusters by UCB value, $O(2*q\log(q))$ with $q$ the size of the largest cluster. It does not depend on the complexity of the CO problem considered, resulting in high-efficiency even for NP-hard and large-scale problems. %Note that after training it is necessary to solve the CO problem to synthesize a desirable solution. 
The computation time of the solutions is amortized in the pre-computation of a large pool of feasible or relaxed solutions, which can easily be parallelized.
%Finally, part of the computation time is offline -- solution pool generation and clustering --, for even faster query selection.

\subsection{Weight Update with Maximum Likelihood}
At each iteration, weight estimates are updated according to the received feedback on the selected query. If the DM is not indifferent, they select one preferred solution, denoted $y_+$ (the other one is denoted $y_-$). In case of indifference, a new query is selected. Existing CPE approaches~\cite{dragone2018constructive,dragone2018hybrid} update preference weights using the Preference Perceptron (PP)~\cite{shivaswamy2015coactive}, a variant of the Structured Perceptron (SP)~\cite{collins2002SP} for preference data. The weight update rule of PP is based on how the two selected solutions differ in sub-objectives, measured by $\Delta = \phi(y_+,p)-\phi(y_-,p)$. In the following, we omit the context $p$ to simplify notation. The PP weight update rule is
\begin{equation*}
    w^{t+1}_\textrm{PP} = w^t_\textrm{PP} + \eta.\Delta
\end{equation*}
with $\eta$ the learning rate.
The original SP is updated only in case of misprediction. As with preference data, the prediction is incorrect if $u^t(y_+) < u^t(y_-)$, the SP-based update rule is:
\begin{equation*}
    w^{t+1}_\textrm{SP} = w^t_\textrm{SP} + \eta. \mathds{1}_{u^t(y_+) < u^t(y_-)}\Delta
\end{equation*}

\subsubsection{Maximum Likelihood}
Instead of considering the learning task as structured-output prediction, we formulate it with Maximum Likelihood Estimation (MLE). Under the well-established Bradley-Terry model~\cite{bradley1952rank}, the probability of $y_+$ being preferred over $y_-$ is 
\begin{equation*}
    P(y_+ \succ y_-) = \frac{e^{u^*(y_+)}}{e^{u^*(y_+)} + e^{u^*(y_-)}}
\end{equation*}
with $u^*$ the true (unknown) DM's utility.
Training with MLE aims to maximize this probability. 
As usual, MLE is re-written as a loss L minimizing the negative log-likelihood: %$L=-\log \frac{e^{u^t(y_+)}}{e^{u^t(y_+)} + e^{u^t(y_-)}} $. 
\begin{align*}
\label{eq:loss}
 L &= - \log \frac{e^{\sum_i w_i \phi_i(y_+)}}{e^{\sum_i w_i \phi_i(y_+)} + e^{\sum_i w_i \phi_i(y_-)}}  \\
 &= -\log \frac{1}{1+e^{-\sum_i w_i [\phi_i(y_+) -\phi_i(y_-)]}} \nonumber \\
 &= -\log \mathcal{S} (\sum_i w_i \Delta_i)   \nonumber
\end{align*}
With $\mathcal{S}$ the sigmoid function and $\Delta_i = \phi_i(y_+) -\phi_i(y_-)$. 
This loss is the standard ML objective to fit a Bradley-Terry model~\cite{rafailov2024dpo}, but has not been used for CPE so far. 

Using gradient descent, the update rule is: $w_{t+1} = w_t - \eta \nabla L$. %with $\eta$ the learning rate. 
The partial gradients w.r.t. to a given weight $w_i$ are: 
$$ \frac{\partial L}{\partial w_i} 
= -\Delta_i \frac{e^{-\sum_j w_j \Delta_j}}{1 +  e^{-\sum_j w_j \Delta_j}} 
= -\Delta_i \frac{1}{1 +  e^{\sum_j w_j \Delta_j}} 
= -\alpha \Delta_i$$
With $\alpha = \mathcal{S}(-\sum_j w_j \Delta_j) = \mathcal{S}(u(y_-) - u(y_+))$. Note that $\alpha$ is independent of $i$ and is therefore the same for all weights $w_i$.
Therefore, the MLE update rule is $$w^{t+1}_\textrm{MLE} = w^t_\textrm{MLE} + \eta. \mathcal{S}(u^t(y_-) - u^t(y_+)) \Delta$$
When an ensemble is used for query selection, this update is applied to each weight vector of the ensemble.

\subsubsection{A unified view}
\begin{figure}[t]
    \centering\includegraphics[width=0.8\linewidth]{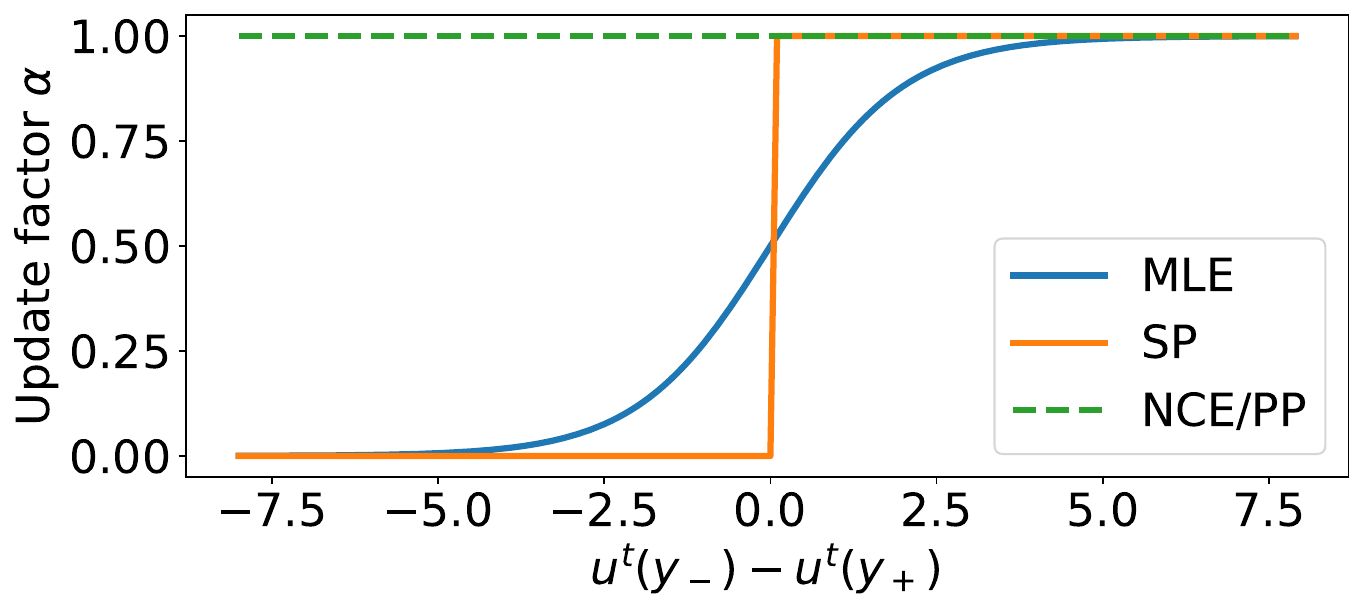}
    \caption{Comparing the update factor of MLE, SP and PP/NCE.}
    \label{fig:update}
\end{figure}
The PP update can be interpreted in the context of Noise Contrastive Estimation (NCE). As in~\cite{mulamba2021contrastive}, we write the probability of a solution $y$ as $P(y|w)=e^{u(y,w)}$. Then, the infoNCE loss~\cite{oord2018infoNCE} can be used to increase the separation between the preferred solution $y_+$ (positive sample) and the other solution $y_-$ (negative sample): $L^{\textrm{NCE}}:= - \log \frac{P(y_+|w)}{P(y_-|w)} 
= - \log \frac{e^{u(y_+, w)}}{e^{u(y_-, w)}}
= - \sum_i w_i \Delta_i
$. 
The partial gradients $\frac{\partial L}{\partial w_i} = - \Delta_i$ lead to the same gradient update as PP: $w^{t+1}_{NCE} = w^t_{NCE} + \Delta$.

All three weights updates -- SP, PP/NCE and MLE -- can be expressed using an update factor $\alpha$ and a learning rate $\eta$: 
\begin{equation*}
w^{t+1} = w^t + \eta. \alpha^t \Delta
\end{equation*}
There is $\alpha^t=\mathds{1}_{u^t(y_+) < u^t(y_-)}$ for SP, $\alpha^t=1$ for PP/NCE and $\alpha^t=\mathcal{S}(u^t(y_-) - u^t(y_+))$ for MLE.
The update factor $\alpha^t$ ranges between $0$ and $1$ and controls the amplitude of the gradient step amplitude.
As plotted in Figure~\ref{fig:update}, MLE can be interpreted as a smoothed version of SP. 
When the difference in predicted utility between $y_+$ and $y_-$ is large, \textit{i.e.,} a clear preference is predicted, the update factor tends to $0$ if the prediction is correct and is maximum ($1$) otherwise. When no clear preference is predicted %(small utility difference)
, the MLE update factor is a function of the utility difference.

\paragraph{Batch learning.}
Previous CPE approaches update weights online, \textit{i.e.,} after each received feedback. We instead propose to train MLE with batch training. We create a feedback dataset and retrain the whole model at each iteration (line~\ref{alg:update} in Algorithm~\ref{alg:CPE}). Since the model is a single linear vector -- the estimated weights $w$ -- fully retraining it is still fast.
%This opens the door to weight prediction from a feature vector (\textit{e.g,} representing a DM, as preferences may vary from one to another) and training neural networks. We leave it for future work.

%%%%%%%%%%%%%%%%%%%% Results %%%%%%%%%%%%%%%%%%%%%%%%%%%%%%%
%%%%%%%%%%%%%%%%%%%%%%%%%%%%%%%%%%%%%%%%%%%%%%%%%%%%%%%%%%%%

\section{Experiments}

%The overall goal is to improve interaction with the DM, which lead to research questions:
%\begin{itemize}
 %   \item[Q1] How does the proposed query selection change the quality/efficiency trade-off.
  %  \item[Q2] How do the different weight updates impact the quality of the synthesized solution (in terms of regret)? %the number of queries?
  % \item[Q3] How does the proposed method generalizes to multiple instances?
%\end{itemize}

\subsection{Experimental Setting}
\paragraph{Baselines.} We re-implemented the two CPE methods applicable for pairwise queries. The first, Set-wise Max-margin (SetMargin)~\cite{teso2016constructive} does not follow the preference elicitation steps but rather turns each data point into a constraint, and the weight vector is optimized to have the largest separation margin. This method is limited to Boolean sub-objectives. The second, Choice Perceptron~\cite{dragone2018hybrid}, was proposed to overcome this limitation and handles both Boolean and numerical sub-objectives.  
With the Choice Perceptron, weights are updated according to the PP weight update, and queries $(y_1, y_2)$ are selected by solving %the following CO problem: 
%\begin{align*}
%    y_1 &= \arg\max_{y\in \mathcal{F}}u^t(y) \\
%    y_2 &= \arg\max_{y\in \mathcal{F}} ((1-\gamma)u^t(y) + \gamma||\phi(y_1)-\phi(y_2)||_1
%\end{align*}
%with $\gamma=1/t$. It maximizes a trade-off between the utility estimate and the diversity (in sub-objectives) between selected solutions. 
as explained in Section~\ref{sec:query}.
%In the following, we refer to this query selection scheme as Feasible + L1. 
It uses $\gamma=1/t$ with $t$ an iteration counter. It also includes costly learning-rate tuning between iterations that we omit for more equal comparison.

\paragraph{Tasks.} We first consider the single-instance task of PC Configuration used by both baselines as a constructive variant of PC recommendation~\cite{guo2010real}. Each configuration is described by $7$ components -- such as storage and CPU model --  and the price. Sub-objectives are Boolean ($77$ in total) indicating if a component is selected, plus the normalized price. We also consider the more realistic multi-instance task of Prize-Collecting Traveling Salesperson Problem (PC-TSP), previously used for structured prediction~\cite{vejar2024efficient}. 
Each node represents a city, which offers a prize if visited and costs a penalty $\rho$ if not. Each edge $(i,j)$ has $k=4$ values $v_k^{ij}$ that can be interpreted as distances, duration, fuel consumption and driver familiarity. The total penalty is a $5$th sub-objective. Mathematically, the CO problem objective is
\begin{equation*}
\min_{y \in {\cal F}} \sum_{l=1}^k w_l \sum_{(i,j) \in E} v_l^{ij} y_{ij} + w_{k+1} \sum_{i \in V} \rho^i (1-\sigma_i)
\end{equation*}
where Boolean variables $y_{i,j}$ (resp. $\sigma_i$) indicate whether the edge $(i,j)$ (resp. city $i$) is part of the solution. The constraints state that the chosen edges form a Hamiltonion circuit. All sub-objectives are normalized by dividing the columns of the distance matrices by its largest value, ensuring all entries lies in $[0, 10]$. % while preserving relative distances. 
We consider problems with 10, 20 and 100 nodes. These problems are multi-instance: we train on $50$ instances and test (\textit{i.e.,} synthesize solutions) on $10$ unseen instances.

\paragraph{User simulation.} We replicate the same experimental protocol as baselines. Each experiment is repeated $20$ times, each with a different DM whose ground-truth utility is obtained by randomly sampling each weight.
For the configuration task, DM weights were sampled from a normal distribution $\mathcal{N}(25, 25/3)$~\cite{teso2016constructive} then $80\%$ were randomly set to $0$ to create sparse preferences. 
%in Guo, draw mu in [0, 100] and std is mu/3
For the PC-TSP, DM weights were sampled from a Dirichlet's distribution of concentration parameter $100$~\cite{vejar2024efficient}. 

The DM's response to a query is simulated based on the Bradley-Terry model extended to indifference, to create noisy responses including indifference~\cite{guo2010real}. 
For both tasks, the quality of synthesized solutions is assessed using relative regret: $\frac{u^*(y^*) - u^*(\hat y)}{u^*(y^*)}$, with $\hat y$ the synthesized solution and $y^*$ the true optimal solution. At test time, we measure how many of the $20$ DMs are satisfied, meaning that they are indifferent between $\hat y$ and $y^*$. % according to the simulation.
We also measure the number of queries needed to reach an average relative regret below 10\%. This threshold is considered an `acceptable' solution by Teso et al~\shortcite{teso2016constructive}.

\paragraph{Implementation.}  The experiments are implemented using CPMpy~\cite{guns2019cpmpy} and GurobiPy~\cite{gurobi}. %device?
MLE was trained both online and with batch training, the latter for $4$ epochs and a batch size of $4$. 
Parameters were tuned based on a separate set of $10$ DMs. 
After tuning, the learning rate for the configuration task is $2$; for PC-TSP it is $0.5$ for SP-online and MLE-online, $0.1$ for PP-online and $1$ for MLE-batch. 
We applied $100$ training steps like~\cite{teso2016constructive,dragone2018hybrid}.
For the ensemble-based acquisition functions, an ensemble of $m=25$ models is used.
%To compute the UCB, $\gamma$ is set to $1/t$ like in~\cite{dragone2018hybrid}. 
Since the sub-objectives are Boolean for the Configuration problem, all configurations with the same number of identical components are equidistant, so no clustering is applied. For PC-TSP, the number of clusters used was $5$. 
For each problem, a pool of $10,000$ different random relaxed solutions was generated, by randomly selecting attributes (configuration task) or by randomly sampling node permutations (PC-TSP). For PC-TSP, varying-size circuits were generated (the circuit is completed when the warehouse node is reached). The code is available at: \url{https://github.com/mdefresne/CPE_pool}.
%as soon as the warehouse node was reached in the permutation, the circuit was completed even though all nodes were not visited.

%\paragraph{Research questions.} Our experiment study focuses on: % 4 research questions:
%\begin{itemize}
 %   \item[RQ1] Runtime of query generation methods
 %   \item[RQ2] Impact of query selection and weight update choice
 %   \item[RQ3] Ablation on solution pool design
 %   \item[RQ4] Comparison to state-of-the-art
%\end{itemize}

%for table, number should be aligned right (for readability)
\begin{table}[t]
    \small
    \centering
    \begin{tabular}{c|rrr}
         Query & SetMargin & ChoicePerc & r-pool (ours) \\
         \midrule
        Configuration & 1.9 & 0.54 & 0.004  \\
        PC-TSP 10 & \textit{NA} & 0.24 & 0.001  \\
        PC-TSP 20 & \textit{NA} & 52.2 & 0.004  \\
        PC-TSP 100 & \textit{NA} & \textit{Timeout} & 0.050  \\
    \end{tabular}
    \caption{Average time (s) for selecting a single query single. SetMargin is not applicable to the PC-TSP (indicated by \textit{NA}).}
    \label{tab:time}
\end{table}
%depends a bit on the number of sobj and size (for dot product)

\subsection{Query Selection Time}
%Our proposed method ensures real-time queries, even for large NP-hard CO problems, by precomputing a pool of solutions before training, then selecting queries within it. 
Table~\ref{tab:time} compares the average time to select a single query for the SetMargin method, the Choice Perceptron method, and our method with a solution pool of $10,000$ relaxed solutions (r-pool). %; on the configuration problem and increasing sizes of prize-collecting TSPs.
Due to its Boolean attribute restriction, SetMargin can only be run on the Configuration problem, and we can see that it has the slowest query selection here. For the Choice Perceptron, we can see that the runtime is heavily influenced by the difficulty of the CO problem, even timing out ($300$s) for PC-TSP problems of size 100. Our query selection using the solution pool takes marginal compute time.

%If both baselines can be considered real-time of the configuration task, on the NP-hard task of PC-TSP the query selection time quickly exploded as problem size increases. Our pool-based query selection, however, is agnostic to the complexity of the underlying CO problem. We indeed observed the time to select one query was independent of the problem and only linearly affected by its size. After training, the CO problem still needs to be solved to synthesize a desirable solution.

Most of our computation time is offline, before interacting with the DM. The generation of the relaxed solution pool takes fewer than $20$ minutes for the largest considered problem (PC-TSP with 100 nodes). 
%generation time for PC-TSP with 100 nodes: 1118s (18.6 minutes)
Note that in the Choice Perceptron, one could also replace the solver call with an argmax call over our solution pool to achieve similar speed-ups.

\subsection{Impact of Query Selection and Weight Update}
We now compare the baseline query selection technique of the Choice Perceptron, with our proposed UCB selection over a pool of relaxed solutions. For each, we also compare the three weight updates discussed (the standard Choice Perceptron uses 'PP'), plus batched MLE. %To distinguish the impact of the MLE update from batch training, an online version of MLE is trained -- it is equivalent to a smooth SP. 
Results for Configuration and for the Prize-collecting TSP are shown in Table~\ref{tab:reg}. %PC-TSP test assesses the generalization to $10$ unseen instances.

Selecting queries with UCB on a relaxed pool outperforms the Choice Perceptron query selection on both tasks and for each weight update except PP + ChoicePerc on Configuration. On the other cases, using UCB leads synthesized solutions of higher quality -- both in regret and %the number of satisfied DMs (
\% DM --
and up to half the number of queries are required. The results are especially pronounced on the multi-instance PC-TSP problem, where using the UCB query selection leads to DMs being satisfied across all the weight updates, and with considerably fewer queries needed to reach 10\% regret (\#Q).

Looking closer at the different weight updates, SP performs the worst overall. The smooth variant MLE-online improves all metrics on both problems and both query selections, showing that the smooth weight update is beneficial. Moving from online to batch re-training (MLE-batch) provides further improvements, in particular in the required number of queries.
Overall, the choice of the UCB query selection with the MLE-batch update rule leads to higher-quality solutions by a wide margin, while requiring fewer queries. %Only on the PC-TSP does the Choice Perceptron with MLE-batch come close.

\begin{table*}[t]
    \small
    \centering
    \begin{tabular}{lrrrrrrrrrrrr}
         & \multicolumn{6}{c}{\textbf{Configuration}} & \multicolumn{6}{c}{\textbf{Prize-Collecting TSP}} \\
         \cmidrule(lr){2-7}
         \cmidrule(lr){8-13}
         Query & \multicolumn{3}{c}{ChoicePerc} & \multicolumn{3}{c}{UCB, rpool (ours)} & \multicolumn{3}{c}{ChoicePerc} & \multicolumn{3}{c}{UCB, rpool (ours)}   \\
         \cmidrule(lr){2-4}
         \cmidrule(lr){5-7}
         \cmidrule(lr){8-10}
         \cmidrule(lr){11-13}
          Update & Regret (\%) & \#Q & \% DM & Regret (\%) & \#Q & \% DM & Regret (\%) & \#Q & \% DM & Regret (\%) & \#Q & \% DM \\
         \midrule
         PP-online  & 6.6 $\pm$ 0.5 & 51 & 55 & 7.7 $\pm$ 0.5 & 67 & 45  & 2.7 $\pm$ 0.2 & 66 & 75 & 1.1 $\pm$ 0.05 & 29 & \textbf{100} \\
         SP-online  & 14.6 $\pm$ 0.7 & - & 20 & 8.9 $\pm$ 0.5 & 69 & 55 & 12.1 $\pm$ 0.7 & - & 40 & 1.2 $\pm$ 0.06  & 22 & \textbf{100} \\
         MLE-online & 8.2 $\pm$ 0.6 & 90  & 65 & 4.9 $\pm$ 0.6 & 70 & 60 & 1.7 $\pm$ 0.1 & 39 & 90 & 0.6 $\pm$ 0.03 & 24 & \textbf{100}\\
         MLE-batch  & 8.6 $\pm$ 0.6 & \textbf{47} & 50 & \textbf{2.2 $\pm$ 0.2} & 50 & \textbf{80} & 1.4 $\pm$ 0.1 & 25 & \textbf{100} & \textbf{0.5 $\pm$ 0.02} & \textbf{20} & \textbf{100} \\

    \end{tabular}
    \caption{PC Configuration and Prize-Collecting TSP (10 stops) for changing query selection (columns) and weight update (rows). \\
    \#Q $(\downarrow)$ is the number of queries to reach 10\% regret, \% DM $(\uparrow)$ is the percentage of satisfied DMs. For regret $(\downarrow)$, standard errors are given.}
    \label{tab:reg}
\end{table*}

\subsection{Ablations}
\paragraph{Acquisition function.}
%We select queries with UCB to balance uncertainty and solution quality.
%, with clustering to ensure diversity.
We assess how restricting the acquisition function to either uncertainty or solution quality affects performances in Table~\ref{tab:relax}. 
% Using only solution quality means UCB is restricted to the ensemble mean (\textit{i.e.,}$\gamma=0$); 
Focusing only on solution quality is achieved by setting $\gamma$ to zero;
it is close to CO-based query selection schemes. Using only uncertainty restricts UCB to the variance ($\gamma=1$) and this is how ensemble-based Active Learning is usually used for single samples. In both cases, the synthesized solution quality drops while requiring more queries. Using only the mean results in $3$ times lower regret than with variance only, confirming the importance of solution quality in a constructive setting.
%We explored the use of time-dependent trade-off $\gamma$ to favour exploration (uncertainty) at early training stages and exploitation (solution quality) at later stages, but this did not show clear improvements (see Appendix xxx).

We also verify the effect of using clustering to increase diversity (only applicable to PC-TSP). We here describe the results of that experiment: when no clustering was used, $98.3$\% of selected pairs were similar, \textit{i.e.,} they would have belonged to the same cluster. This lack of diversity led to more indifferent answers: $8.0$\%, vs $0.7$\% with clustering. The quality of synthesized solutions also dropped, with a regret of $13.5$\% (vs $0.5$\%) and only $40\%$ (vs 100\%) of DMs were satisfied. % (vs $100\%$ with clustering).
%End regret: 13.526 +- 0.8333999999999999, Within 10% of best utility in -1 queries, 40.0% of satisfied users

\paragraph{Feasible vs relaxed solutions.}
We compare training on feasible solutions or generating relaxed solutions. On the configuration task, all the feasible solutions ($\sim 60,000$) 
can be enumerated and the results are shown in Table~\ref{tab:relax}.
%Hence, taking the argmax within the pool is equivalent to a solver call, and faster.
% Since the result of a 'solve' call will always be in the feasible pool, the argmax on the pool is equivalent to calling a solver. 
% Training time is then much smaller because done upfront. 
%On the other hand, 
Generating the feasible pool takes about $30$ times longer than randomly generating a pool of $10,000$ relaxed solutions. 
Training on a pool of relaxed solutions (\textbf{r}-pool) instead of feasible ones (\textbf{f}-pool) only marginally degrades metrics. Therefore, a random pool offers a much better efficiency/quality trade-off. 
%We further assessed the impact of the relaxed pool size on Appendix~\ref{app:pool}.

\begin{table}[tb]
    \small
    \centering
    \begin{tabular}{lrrrr}
          & Regret (\%) & \#Q & \% DM & PoolGen (s)\\
        \midrule
        Variance, \textbf{r}-pool & 8.0 $\pm$ 0.5 & 78 & 45 & 0.3 \\
        Mean, \textbf{r}-pool & 2.7 $\pm$ 0.2 & 55 & 60 & 0.3 \\
        \midrule
        UCB, \textbf{r}-pool & 2.2 $\pm$ 0.2 & 50 & 80 & 0.3  \\
        UCB, \textbf{f}-pool & 1.9 $\pm$ 0.2 & 54 & 85 & 9.5 \\
    \end{tabular}
    \caption{PC Configuration, varying acquisition functions, MLE-batch weight update. PoolGen is upfront time for pool generation.}
    \label{tab:relax}
\end{table}

%\begin{figure}[t]
%    \centering
%    \includegraphics[width=0.8\linewidth]{Figures/tSNE_TSP.pdf}
%    \caption{t-SNE representation of the clusters of the PC-TSP sub-objective space.}
%    \label{fig:tSNE}
%\end{figure}

\subsection{Comparison to State-of-the-art}
\begin{table}[b]
    \small
    \centering
    \begin{tabular}{lrrrr}
          & Regret (\%) & \#Q & \% DM & Train (s) \\
         \midrule
        SetMargin & 4.9 $\pm$ 0.7 & 62 & 75 & 348 \\
        Choice Perceptron & 6.6 $\pm$ 0.5 & 51 & 55 & 64 \\
        UCB, r-pool (ours) & \textbf{2.2 $\pm$ 0.2} & \textbf{50} & \textbf{80} & \textbf{4}  \\
    \end{tabular}
    \caption{Our method and the baselines on the configuration task.}
    \label{tab:baseline}
\end{table}
%Training time (without regret computation: SetMargin 348.8 s, Our 4.1s, ChoicePerc 63.9

%\begin{figure}[t]
 %   \centering
  %  \includegraphics[width=\linewidth]{Figures/resultsTSP.pdf}
   % \caption{Relative regret during training on the PC-TSP.}
    %\label{fig:TSP}
%\end{figure}

We compare our complete proposed method, combining our query selection (UCB, r-pool) with the MLE-batch weight update, to previous CPE baselines, SetMargin~\cite{teso2016constructive} and Choice Perceptron~\cite{dragone2018hybrid}.
On the configuration task (Table~\ref{tab:baseline}), our method outperforms both baselines in terms of quality of synthesized solutions while training an order of magnitude faster.

On the PC-TSP with 10 nodes (Table~\ref{tab:reg}), SetMargin is not applicable due to numerical sub-objectives.
Our method outperforms the Choice Perceptron (with its native PP-online weight update) by reaching twice lower regret, requiring half the queries and satisfying $100\%$ of DMs (vs $75\%$). 
We also want to test the approaches on larger instances; however query selection time becomes prohibitive for Choice Perceptron. To enable a comparison, we run both our method and the Choice Perceptron on our randomized pool of solutions for PC-TSP with 20 nodes in Table~\ref{tab:TSP20}.
%As before, $50$ instances are used for training and $10$ other for testing. 
Training on the precomputed pool is fast and leads to high-quality solution with both methods. Our UCB-based acquisition function with the MLE-batch weight update results in half the regret and 10\% more satisfied DMs than (relaxed) Choice Perceptron. Training time per iteration is longer due to batch training, but remains under one second.

\begin{table}[tb]
    \small
    \centering
    \begin{tabular}{lrrr}
          & Regret (\%) & \% DM & Train (s) \\
         \midrule
         Choice Perc, r-pool & 2.9 $\pm$ 0.30 & 85 & 0.44 \\
         UCB, r-pool & \textbf{1.4 $\pm$ 0.06}  & \textbf{95} & 0.93 \\
    \end{tabular}
    \caption{PC-TSP with 20 nodes. Average training time per iteration.}
    \label{tab:TSP20}
\end{table}

\section{Conclusion}
We revisited CPE to learn preferences over the sub-objectives of a multi-objective CO problem. We targeted desirable properties for preference elicitation systems: real-time, robust to noisy answers and a minimal number of queries while synthesizing high-quality solutions.
Previous solutions repeatedly called a CO solver %to generate solutions 
during query selection; 
%All previous methods selected feasible solutions, requiring repeatedly solving CO problems. W
we instead 
% built a pool of relaxed solutions from which queries are selected 
select from a pool of relaxed solutions, ensuring \emph{real-time} query selection even for NP-hard problems, which in turn enabled us to use ensemble-based activation functions.
%bypassing the need of solving CO problem during training
We also proposed the smoothed MLE weight update, and a batch retraining strategy. In the experiments, our method synthesized higher-quality solutions, both in terms of regret and DM satisfaction, and required fewer queries than the baselines. %It also generalized the learnt preferences to unseen instances. 

Multiple avenues for further methodological developments are possible. First, we sampled solution pools using problem-specific relaxations. Automating the relaxation for any CO problem would broaden the applicability of this approach; as would techniques for sampling feasible solutions. The challenge there is to obtain a diverse and unbiased sampling in an efficient way~\cite{sampling_csp}.
We ran all experiments for 100 iterations as in previous work; developing an instance-dependent stopping criterion would be desirable as it could restrict the number of needed DM interactions or continue to obtain even better performance.
An acquisition function defined over pairs directly could potentially further boost efficiency.
And while we used simple linear models, the weight updates can be used in gradient descent for arbitrary neural networks; meaning DM-specific features could also be taken into account. 
%our proposed weight update enables predicting weights from any features and training by gradient descent, with neural network for instance. It could be used to adapt preferences to different DMs.
%sample a pair, not 2 solutions independently
Finally, by improving on the desirable properties, we hope our method opens the door to real-life testing on rich multi-objective optimization problems.

\clearpage
\appendix

%\section*{Ethical Statement} % Optional
%There are no ethical issues.

\section*{Acknowledgments} % In camera-ready
This research received funding from the European Research Council (ERC) under the EU Horizon 2020 research and innovation programme (Grant No 101002802, CHAT-Opt). Jayanta Mandi is supported by the Research Foundation-Flanders (FWO) project G0G3220N. 

\bibliographystyle{named}
\bibliography{ijcai25}

\end{document}